# Artificial Intelligence Hybrid-Deep Learning Model for Groundwater Level Prediction Using MLP-ADAM

Pejman Zarafshan, Saman Javadi, Abbas Roozbahani, Seyed Mehdi Hashemy
Department of Irrigation and Drainage Engineering, College of Aburaihan, University of Tehran
Tehran, Iran

Payam Zarafshan, Hamed Etezadi
Department of Agro-Technology, College of Aburaihan, University of Tehran
Tehran, Iran
p.zarafshan@ut.ac.ir

*Abstract*— Groundwater is the largest storage of freshwater resources, which serves as the major inventory for most of the human consumption through agriculture, industrial, and domestic water supply. In the fields of hydrological, some researchers applied a neural network to forecast rainfall intensity in space-time and introduced the advantages of neural networks compared to numerical models. Then, many researches have been conducted applying data-driven models. Some of them extended an Artificial Neural Networks (ANNs) model to forecast groundwater level in semi-confined glacial sand and gravel aquifer under variable state, pumping extraction and climate conditions with significant accuracy. In this paper, a multi-layer perceptron is applied to simulate groundwater level. The adaptive moment estimation optimization algorithm is also used to this matter. The root mean squared error, mean absolute error, mean squared error and the coefficient of determination ($R^2$) are used to evaluate the accuracy of the simulated groundwater level. Total value of $R^2$ and RMSE are 0.9458 and 0.7313 respectively which are obtained from the model output. Results indicate that deep learning algorithms can demonstrate a high accuracy prediction. Although the optimization of parameters is insignificant in numbers, but due to the value of time in modelling setup, it is highly recommended to apply an optimization algorithm in modelling.

## I. Introduction

Groundwater is the largest storage of freshwater resources, which serves as the major inventory for most of the human consumption through agriculture, industrial, and domestic water supply, [1-2]. With the growth of computer science and hardware, the data-driven methods (e.g., regression analysis, statistics, gray theory, machine learning) have been extensively studied and broadly used in lots of areas with good-enough results, [3-5]. In the fields of hydrological, reference [6] applied a neural network to forecast rainfall intensity in space-time and introduced the advantages of neural networks compared to numerical models. Then, many researches have been conducted applying data-driven models. Also, an Artificial Neural Networks (ANNs) model is extended to forecast groundwater level in semi-confined glacial sand and gravel aquifer under variable state, pumping extraction and climate conditions with significant accuracy, [7]. Reference [8] found Feedforward Neural Network (FNN) models to predict monthly and quarterly time-series water levels at a monitoring well with higher results than the models developed by statistical regression methods. Moreover, the capabilities of Auto-Regressive Moving Average, Adaptive Neuro-Fuzzy Inference System, ANN, Support Vector Machine and Gene Expression Programming are investigated for groundwater level forecast, [9]. Reference [10] used hybrid models by synthesizing Ensemble Empirical Mode Decomposition (EEMD) with data-driven models to forecast groundwater level fluctuations. Thus, Deep Learning (DL) algorithms have shown high potential in areas such as speech recognition, computer vision, natural language processing and intelligent recommendation. Reference [10] applied deep learning algorithms to enhance simulations of large-scale groundwater flow in IoTs. Also, groundwater storage is estimated from seismic data using deep learning, [11]. Reference [12] used Multi-Layer Perceptron and Long Short-Term Memory beside with one statistical regression model, Seasonal Autoregressive Integrated Moving Average to predict the model. Three approaches are compared for optimizing the models' hyper parameters, including two surrogate model-based algorithms and also a random sampling method, [13]. Results indicated that the optimization of the hyper parameters leads to the reasonably more accurate performance of all models and forecast the groundwater level. Reference [14] used deep learning to predict groundwater depth fluctuations. As founded on this research, the deep learning model is an intelligent tool to forecast groundwater depths. In addition, advanced AI techniques can preserve resources and labor conventionally employed to estimate different features of complex groundwater systems. Multilayer

Perceptron is used with Particle Swarm Optimization to Predict Landslide Susceptibility, [15]. The results demonstrated that the MLP-PSO model had the most accuracy in prediction. Also, the combination of MLP-PSO algorithm optimization increased the accuracy compared to the MLP model. Reference [16] applied Long Short-Term Memory (LSTM) deep neural network to reconstruct missing groundwater level data and also increase the time series length. An appropriate methodology is exerted to predict the water flow and water level parameters of the Shannon River in the long term based on deep learning. They employed a machine learning model based on deep convolutional neural networks (CNN). The proposed model predicted water flow and also water level from 2013–2080, [17]. Reference [18] studied the feasibility of executing deep learning emulators for contaminant transport models and a generalized flow based on the hydrogeology of an aquifer in South Australia. A data-driven model is employed to forecast groundwater levels in an arid basin using Artificial Neuron Networks (ANNs), Support Vector Machines (SVMs), and M5 Model Tree, [19]. Reference [20] compared nonlinear autoregressive networks with exogenous inputs (NARX) with popular state-of-the-art Deep Learning techniques such as long short-term memory (LSTM) and convolutional neural networks (CNN) to predict groundwater level. A multilayer perceptron (MLP) model, a Seasonal Autoregressive Integrated Moving Average (SARIMA) model, and long short-term memory (LSTM) model are compared to predict slope movements. Results indicated that the moving-average models (SARIMA) performed better than the Machine Learning models (MLP and LSTM) during both training and test, [21].

In this study, a Multi-Layer Perceptron (MLP) is applied to simulate groundwater level. Also, the Adaptive moment estimation (Adam) optimization algorithm is used to this matter. Afterward, this hybrid optimal approach is applied to forecast the groundwater level in the Najafabad aquifer. Finally, to be precise about the accuracy, a comprehensive study of the presented model is presented and the obtained results are discussed.

## II. METHODOLOGY

### A. Case study

The case study aquifer is located in Najafabad plain, Gavkhoni catchment consists of 21 basins. Najafabad study area is located in the middle of this basin and in Isfahan province. The total area of this area is 1754.9 $km^2$, of which 679.2 $km^2$ are highlands and 1075.7 $km^2$ are plains. The alluvial aquifer is spread over an area of 941 $km^2$ (95% of the plain area). The highest point with an altitude of 2953 $m$ is in the northwest and the minimum elevation of 2069 $m$ above sea level is in the western north near the Zayandehrood River. Zayandehrood River enters from the southern part of this area and exits from the eastern part. The water table depth maps of the study region show a dominant northeast-southwest inclination. In the central parts of the plain, the high density of wells created bulls-eye contours and a cone of depression in water table maps (Figure 1). At the southwestern limits of the aquifer, contours align more or less in a northeast-southwest direction. The Najafabad sub-basin has a mainly semiarid climate. The average rainfall is only 155 $mm/year$, and most of the rainfall takes place in the winter from December to April. While during the summer, there is no effective rainfall. Furthermore, annual potential evapotranspiration is about 1,950 $mm$.

### B. Modelling setup

First, by considering the hydrological conditions of the region, the effective parameters of the groundwater table are identified. Since the study area is one of the semi-arid regions of the country, according to studies, there is no correlation between data such as humidity, solar coefficients and other meteorological data with the groundwater table. Also, due to the seasonality of rivers in the region, river discharge data and infiltration are not considered. Finally, rainfall, temperature and water inputs are used to measure, model and predict the water table in the study area of Najafabad aquifer and aquifers with the same conditions. Modelling with this data is performed in 9 steps (Figure 2). After collecting the necessary inputs for modelling from regional water in Isfahan, the data have been sorted.

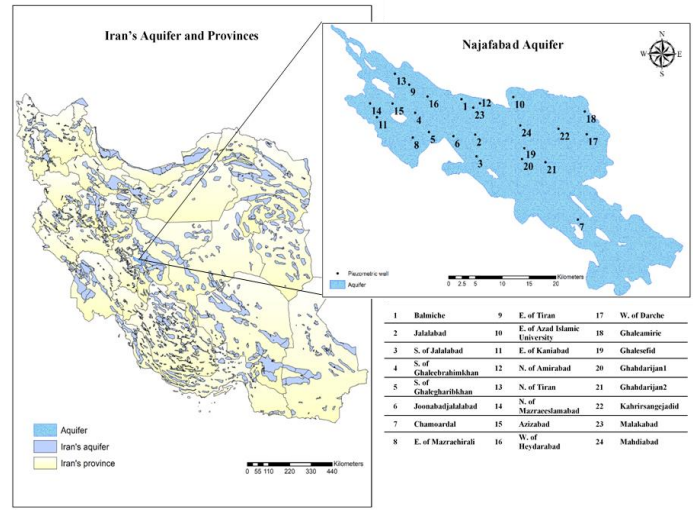

Fig. 1. Study area

### C. Multi-Layer Perceptron (MLP) model

Multi-Layer Perceptron (MLP) model is a feed-forward neural network that consists of an input layer, at least one hidden layer, and an output layer, [22]. It can model nonlinear patterns by introducing multiple layers and nonlinear activation functions that transform input signals. It has been primarily used for classification and function estimation problems, [23]. MLPs have been used for, e.g., forecasting drought, [24], estimating evapotranspiration, [25], and predicting water flow, [26]. Unlike other statistical models, the MLP does not make any assumptions regarding the prior probability density distribution of the training data, [27].

In an MLP network, all the neurons in the hidden and output layers use nonlinear activation functions to simulate the action potential of biological neurons. So, a Rectified Linear Unit (ReLU) function is applied for the activation function in all of the neurons using as:

$$\mathrm{Re}LU(x) = \begin{cases} x \text{ if } x > 0 \\ 0 \text{ if } x \leq 0 \end{cases} \quad (1)$$

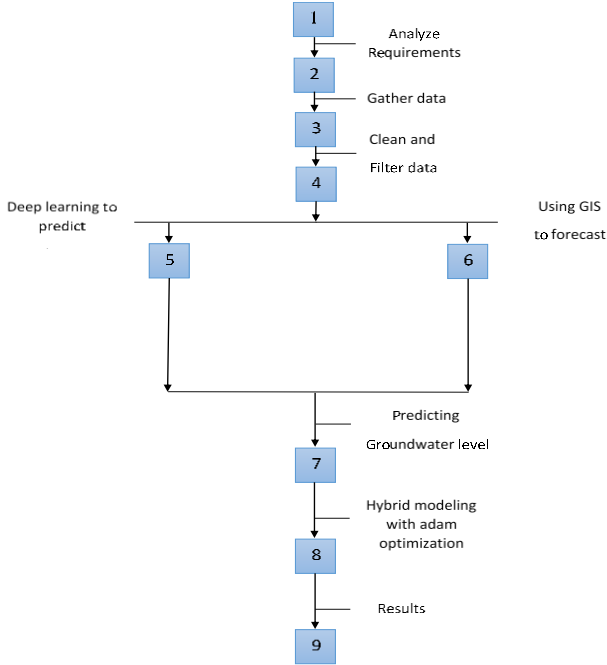

Fig. 2. Research flowchart

Compared with logistic sigmoid function and hyperbolic tan function, the ReLU has finer performance. It can solve the challenges of gradient explosion and gradient disappearance, and preserve the convergence rate in a stable state. Since the network is entirely connected, each node in one layer connects to every node in the next layer with a certain weight. Therefore, the output c of each neuron is:

$$c = \varphi\left(\sum_i w_i a_i + b\right) \quad (2)$$

where $a_i$ and $w_i$ are the inputs and weights of the neuron respectively, b is the bias of the current neuron and φ is the activation function ReLU.

*D. Adaptive Moment Estimation (Adam)*

Adaptive Moment Estimation (Adam) optimization is applied to calculate the adaptive training rate of the parameters, [28]. This method, in addition to storing the descending mean of the square of the past of the gradient or $v_t$ and the mean of the descending of the past of the gradient or $m_t$, is kept as a momentum. This is while the momentum can be seen as a ball sliding on a sloping surface that has no friction and therefore can be placed at the minimum error level, [29]. Both parameters of the mean descending average of the square of the gradient and the average of the descending of the past of the gradient can be calculated as follows:

$$\begin{aligned} m_t &= \beta_1 m_{t-1} + (1-\beta_1) g_t \\ v_t &= \beta_2 v_{t-1} + (1-\beta_2) g_t^2 \end{aligned} \quad (3)$$

$m_t$ and $v_t$ are estimates of the first moment (the mean) and the second moment (the uncentered variance) of the gradients respectively, hence the name of the method. As $m_t$ and $v_t$ are initialized as vectors of zeros, the authors of Adam observe that they are biased towards zero, especially during the initial time steps, and especially when the decay rates are small (i.e. $\beta_1$ and $\beta_2$ are close to 1). They counteract these biases by computing bias-corrected first and second moment estimates:

$$\begin{aligned} \hat{m}_t &= \frac{m_t}{1-\beta_1^t} \\ \hat{v}_t &= \frac{v_t}{1-\beta_2^t} \end{aligned} \quad (4)$$

They then use these to update the parameters just as we have seen in Adadelta and RMSprop, which yields the Adam update rule:

$$\theta_{t+1} = \theta_t - \frac{\eta}{\sqrt{\hat{v}_t} + \varepsilon} \hat{m}_t \quad (5)$$

Default values of 0.9 for $\beta_1$, 0.999 for $\beta_2$, and $10^{-8}$ for ε has been modified. They show empirically that Adam works well in practice and compares favorably to other adaptive learning-method algorithms.

III. RESULTS AND DISCUSSION

The settings of the multi-layer perceptron model are based on the active function of Relu. In this research, network training is based on 80% of available data and testing is based on the remaining 20%. The last column of the model, which represents the groundwater level or output of the model, is shown in Figure (3). The information in this column is obtained from the impact level of each well in the aquifer by a mathematical formula within GIS software. On the other hand, we assume that the inputs of each model include temperature, precipitation and weighted average of groundwater level (in order to reduce the number of features and since the features of reservoirs are one-dimensional, weighted average is used and their output is groundwater level.

In this research, the Python software is used to study the simulations. So, there are many benefits to use a Python environment such as the availability of tables and graphs and many other steps applied in modeling, called libraries. Different libraries are used for various purposes, which will be described. The Pandas library is used to work with *.xls and *.xlsx input formats in modeling. The Numpy, Sklearn, and Tensorflow.keras libraries are employed for the modeling section, and the Matplotlib libraries are exerted to display the output in the form of a graph (Table 1). Adjustment of hyperparameter coefficients is mentioned below. Due to the vast application in various modeling fields, 500 neurons in the hidden layer are applied in the configuration of the multilayer perceptron model (Table 2). Also, the active function of this model is a Rectified linear unit. Moreover, the input layer has 100 nodes, which corresponds to the 100 potential distinguished predictors. There are 500 neurons in the hidden layer. All of the weights and biases of the linkages among the potential variables and groundwater recharge values are improved when the parameters of the network are learned based on the training data. Then the predicted values are computed based on the learned weights and biases when the input data is applied. The output layer only had one neuron, which represented the values of groundwater recharge.

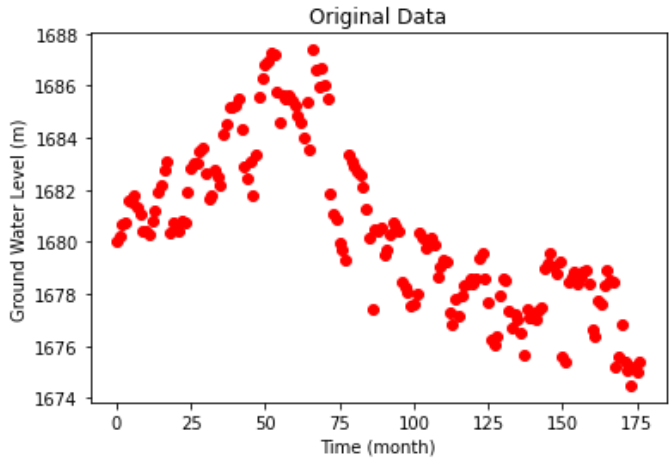

Fig. 3. Water level measured in 175 months

TABLE I. LIBRERIES USED IN THE MLP MODELLING

| Libraries Used | Specifications |
| --- | --- |
| Pandas | Dataframes/importing CSV files |
| Numpy | Scientific Computing |
| Sklearn | Model implementation |
| Matplotlib | Plotting of numerical mathematics |
| Tensorflow.keras | High-level Neural Network APIs |

TABLE II. MULTI-LAYER PERCEPTRON MODEL SETTINGS

| Model summary | Neurons | Activation | Optimizer | inp_size | verbose |
| --- | --- | --- | --- | --- | --- |
| MLP | 500 | Relu | 'adam' | 1 | False |

The results can be studied by comparing the Root Mean Squared Error (RMSE), Mean Absolute Error (MAE), Mean Squared Error (MSE), and coefficient of determination (R2) as shown in Table (3). As stated before, network training is based on 80% of available data, and testing is based on the remaining 20%. The last column of the data, which represents the groundwater level or output of the model, is shown in Figure (3). The information in this column is obtained from the impact level of each well in the aquifer by a mathematical formula within GIS software. On the other hand, the inputs of each model are considered as temperature, precipitation, and weighted average of groundwater level (in order to reduce the number of features, the weighted average is used to obtain the one-dimensional features of wells) and their output is groundwater level. The modeling is performed using a Multi-Layer Perceptron (MLP) model learning simulator in Python. As you can see in Figure (4), the green line represents the MLP model, which is also comparable to the original data.

TABLE III. MODELING OUTPUT OF MULTI-LAYER PERCEPTRON MODEL

| RMSE | MAE | MSE | $R^2$ | Amount | Model |
| --- | --- | --- | --- | --- | --- |
| 0.7308 | 0.5564 | 0.5341 | 0.9436 | 80% | Train |
| 0.7337 | 0.5632 | 0.5383 | 0.9517 | 20% | Test |
| 0.7313 | 0.5578 | 0.5349 | 0.9458 | 100% | Total |

The "Model" column spans MLP for all three rows.

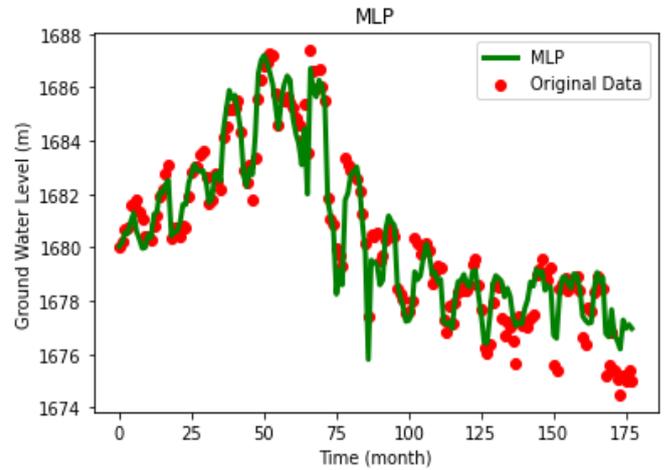

Fig. 4. Model simulation with multi-layer perceptron

IV. CONCLUSION

Modern time-series modeling and forecasts are mostly consisting of artificial intelligence models. In this study, a deep learning structure was conducted to estimate groundwater level. To this end, a Multi-Layer Perceptron (MLP) combining an Adam optimization is exerted. The Root Mean Squared Error (RMSE), Mean Absolute Error (MAE), Mean Squared Error (MSE), and the coefficient of determination ($R^2$) were used to evaluate the accuracy of the simulated groundwater level. The total value of $R^2$ and RMSE are 0.9458 and 0.7313 respectively which were obtained from the model output. Results indicated that deep learning algorithms can demonstrate a high accuracy prediction. Although the optimization of parameters was insignificant in numbers, due to the value of time in modelling setup it is highly recommended to apply an optimization algorithm in modeling.